\documentclass[journal]{IEEEtran}
\usepackage{cite}

\ifCLASSINFOpdf
   \usepackage[pdftex]{graphicx}
   \graphicspath{{../pdf/}{../jpe/}}
   \DeclareGraphicsExtensions{.pdf,.jpg,.png}
\else
\fi
\usepackage{amsmath}

\usepackage{amssymb}
\usepackage{mathtools}
\newcommand{\bigtheta}[1]{\mathit{\Theta}_{#1}}
\usepackage{multirow}
\usepackage{color}

\begin{document}
%
\title{A Multimodal, Enveloping Soft Gripper: Shape Conformation, Bioinspired Adhesion, and Expansion-Driven Suction}
%
%
%

\author{Yufei~Hao, Shantonu~Biswas, Elliot Hawkes, Tianmiao~Wang, Mengjia~Zhu, Li~Wen, and Yon Visell$^{\ast}$
\thanks{Y. Hao, T. Wang and L. Wen are with School of Mechanical Engineering and Automation, Beihang University, Beijing, China.}
\thanks{S. Biswas is with California NanoSystems Institute, University of California, Santa Barbara, CA, USA.}
\thanks{E. Hawkes is with Department of Mechanical Engineering, University of California, Santa Barbara, CA, USA.}
\thanks{M. Zhu and Y. Visell are with Media Arts and Technology Program, Department of Electrical and Computer Engineering, University of California, Santa Barbara, CA, USA. e-mail: yonvisell@ucsb.edu.}} 

%
%

\markboth{IEEE Transactions on Robotics,~Vol.~, No.~, Month~Year}%
{Shell \MakeLowercase{\textit{et al.}}: A Multimodal, Enveloping Soft Gripper: Shape Conformation, Bioinspired Adhesion, and Expansion-Driven Suction}
%



\maketitle

\begin{abstract}
A key challenge in robotics is to create efficient methods for grasping objects with diverse shapes, sizes, poses, and properties.  Grasping with hand-like end effectors often requires careful selection of hand orientation and finger placement.  Here, we present a soft, fingerless gripper capable of efficiently generating multiple grasping modes. It is based on a soft, cylindrical accordion structure containing coupled, parallel fluidic channels.  It is controlled via pressure supplied from a single fluidic port. 
Inflation opens the gripper orifice for enveloping an object, while deflation allows it to produce grasping forces. The interior is patterned with a gecko-like skin that increases friction, enabling the gripper to lift objects weighing up to 20 N.  Our design ensures that fragile objects, such as eggs, can be safely handled, by virtue of a wall buckling mechanism. The gripper can integrate a lip that enables it to form a seal and, upon inflating, to generate suction for lifting objects with flat surfaces. The gripper may also be inflated to expand into an opening or orifice for grasping objects with handles or openings.  We describe the design and fabrication of this device and present an analytical model of its behavior when operated from a single fluidic port.  In experiments, we demonstrate its ability to grasp diverse objects, and show that its performance is well described by our model. Our findings show how a fingerless soft gripper can efficiently perform a variety of grasping operations. Such devices could improve the ability of robotic systems to meet applications in areas of great economic and societal importance.
	
\end{abstract}

\begin{IEEEkeywords}
Shape Conformation, Bioinspired Adhesion, Suction, Soft Gripper 
\end{IEEEkeywords}

%
\IEEEpeerreviewmaketitle

\section{Introduction}
%
%
%
%
\IEEEPARstart{R}{obotic} grasping and manipulation is challenging in many applications, especially those that involve objects with varying sizes, shapes, poses, and properties.  This has motivated the development of a variety of hand-like robotic grippers with multiple fingers  \cite{melchiorri2008robot}.  It has also led to the development of proprioceptive, force, and torque sensors and algorithms and hardware for robotic perception of objects' shape, pose, and properties,  for planning and controlling robotic grasping, and achieving form/force closure  \cite{paolini2014data,pollard2004closure}. Such grasping processes often involve computational scene perception and understanding, or online sensor feedback  \cite{ficuciello2019vision}. The uncertainties arising in practical applications, and the limited compliance of many grippers, make it especially challenging for robotic systems to handle fragile, unfamiliar, or brittle objects.

Recent research on soft robotic grippers has led to several new proposals for improving robotic grasping  \cite{shintake2018soft}. Soft grippers ensure compliant interactions with objects due to their intrinsic compliance.  In the present paper, we use the term ``soft'' to denote objects or systems that are very elastic, as measured by the elastic modulus or other material properties.  Soft robots may be fabricated using techniques that are amenable to multi-material customization, including casting methods based on two component liquid polymers. This has also led to many approaches to actuation.  They include electroactive polymer   \cite{shintake2016versatile,lau2017dielectric,shian2015dielectric}, electromagnetic  \cite{do2018miniature,diller2014three}, thermal  \cite{hubbard2019shrink,breger2015self}, light reaction  \cite{pilz2019untethered}, chemical stimulation  \cite{abdullah2018self,zheng2018programmed}, and fluidic actuation via differential pressure   \cite{mosadegh2014pneumatic,uppalapati2018towards,tawk2018bioinspired}.

 Pneumatic grippers, which are often made of cast silicone elastomers, have been widely investigated because of their low cost, high performance and environmental robustness. They have been deployed in both terrestrial operation \cite{hao2018soft} and underwater sampling \cite{galloway2016soft,yuk2017hydraulic}. More degrees of freedom can be introduced in such grippers through the use of multiple, separated air chambers or by  pre-programming bending locations via functional materials. The latter can improve the  adaptability of a gripper to objects of different sizes and shapes \cite{hao2018eutectic,zhou2018bcl,yang2017bioinspired}. The load capacity of such grippers can be improved by employing variable stiffness structures and mechanisms, including materials of greater or adjustable rigidity  \cite{wei2016novel,jiang2019chain,yap2016high,zhu2019fully,zhang2019fast}. Such improvements can significantly enhance grasping performance.

However, both soft and rigid fingered grippers present challenges in robotic grasping, due to the discrete contacts produced by the fingers and bending mode of actuation. For example, a fingered gripper may generate a reverse moment at the contact point when touching objects, which may push the fingers away from the object. Gaps between the fingers and the target object cannot be eliminated, which may inhibit grasping
 stability. In addition, when grasping an object, a robot using a fingered gripper must account for the positions and orientations of the fingers in relation to the geometry and pose of the object in order to determine a feasible grasping strategy. 

Vacuum-driven, non-fingered grippers, based on granular jamming or the origami designs, can overcome some of these limitations by conforming to objects of arbitrary shapes \cite{brown2010universal,amend2012positive,8794068}. However, such devices can grasp objects from a limited range of orientations. In addition, achieving higher holding forces with such grippers requires greater (negative) pressure, which may pose problems when they manipulate very fragile objects. 

In this article, we present a simple but multi-functional fingerless soft gripper that is able to grasp objects in a variety of poses, sizes, and shapes. The gripper consists of an array of parallel chambers in a soft accordion structure that forms a cylindrical aperture. It is controlled via pressure supplied from a single fluidic port.  With this single control input, the gripper is able to produce three different compliant grasping behaviors.  In the first mode (contraction-based grasping), inflation opens the orifice of the gripper for enveloping an object to be grasped. Subsequently, deflating it envelops the object and produces grasping forces.  The grasping force is improved via a bioinspired, gecko-like patterned skin, increasing the lifting capacity.  The design facilitates a wall buckling mechanism that ensures that  fragile objects, such as eggs, can be safely handled.  In the second grasping mode (expansion-based grasping), the gripper can inflate to expand into an opening or orifice for grasping objects with handles or openings.  The third grasping mode (expansion-driven suction) is enabled by augmenting the gripper with a lip that enables it to form a seal with flat objects.  Subsequently inflating the gripper (and thereby expanding the interior region) produces a suction that enables the gripper to lift objects via flat surfaces. 

The structure of the paper is as follows, we first describe the design and fabrication of this device, and explain how it ensures grasping performance and safety.  We then present an analytical model describing the ways that the gripper responds to pressure inputs supplied via the input port. In a series of experiments, we demonstrate its ability to grasp diverse objects by adapting to their shape, and characterize the gripping forces and performance of the device in each of the three grasping modes.  The contributions of this work include a new method and device for grasping a variety of objects by passively adapting to their shapes, methods for producing several different grasping behaviors via the same soft gripper, and a method of grasping that can simultaneously achieve high load capacity via gecko-like adhesion and safely handling of fragile objects by virtue of a wall buckling property.

\begin{figure*}[t!]
	\centering
	\includegraphics[width=150mm]{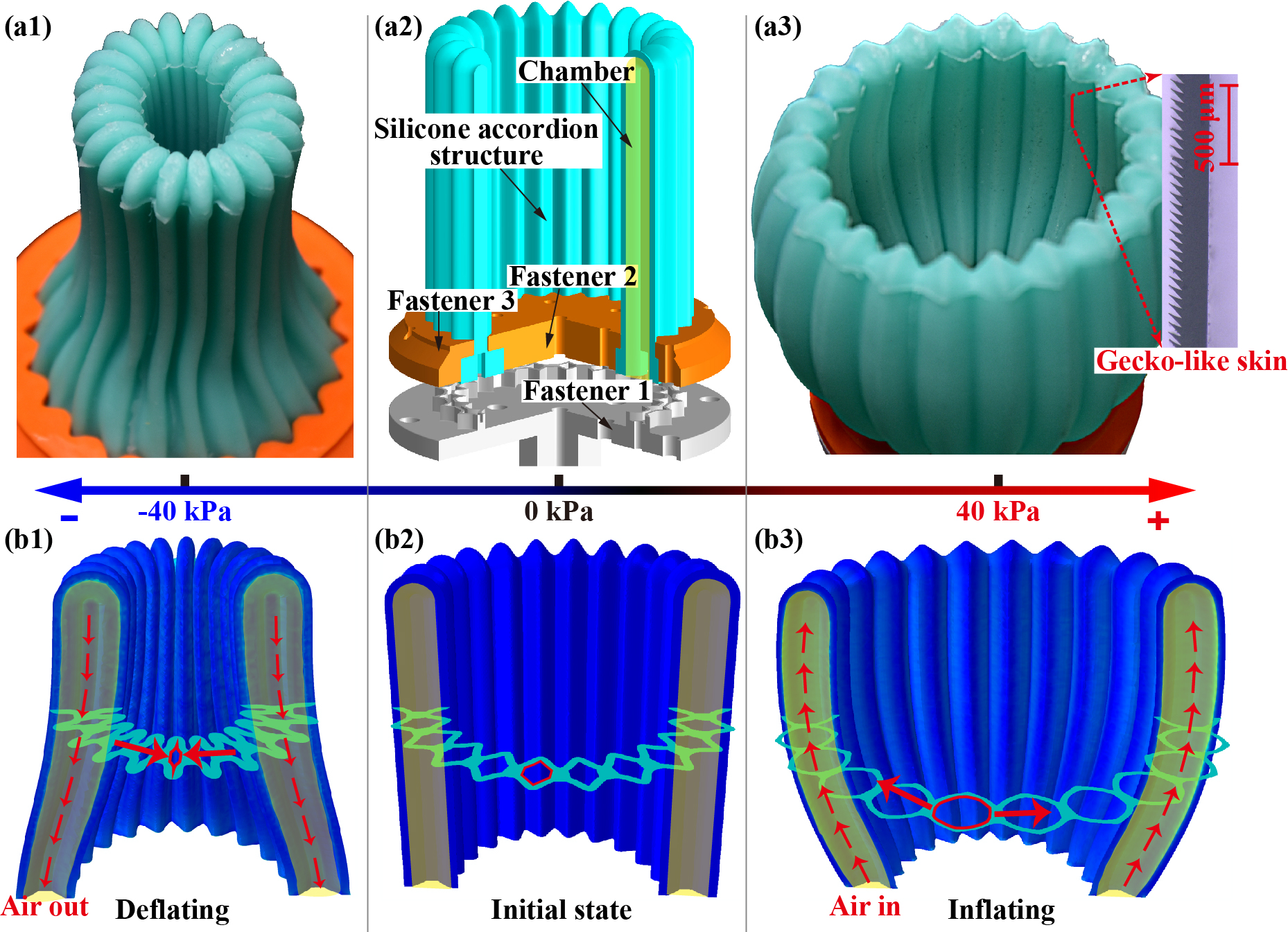}
	\caption{
		 The design and working principle of the gripper. (a1) The gripper contracts as pressure is decreased, closing the aperture. (a2) This is enabled via an array of 22 parallel channels embedded in a silicone accordion structure, all connected to a common fluid port. (a3) Inflating the gripper causes it to expand, exposing the gecko-like skin covering the interior region. (b1) At negative differential pressures, contraction is produced via folding of the soft accordion structure. (b2) The state of the gripper without any input pressure change.  (b3) The expansion of the gripper is produced through positive pressure supplied to the channels, which causes the accordion structure to spread. 
	}
	\label{fig:design_and_ principal} 
\end{figure*}
\begin{figure}[!ht]
	\centering
	\includegraphics[width=75mm]{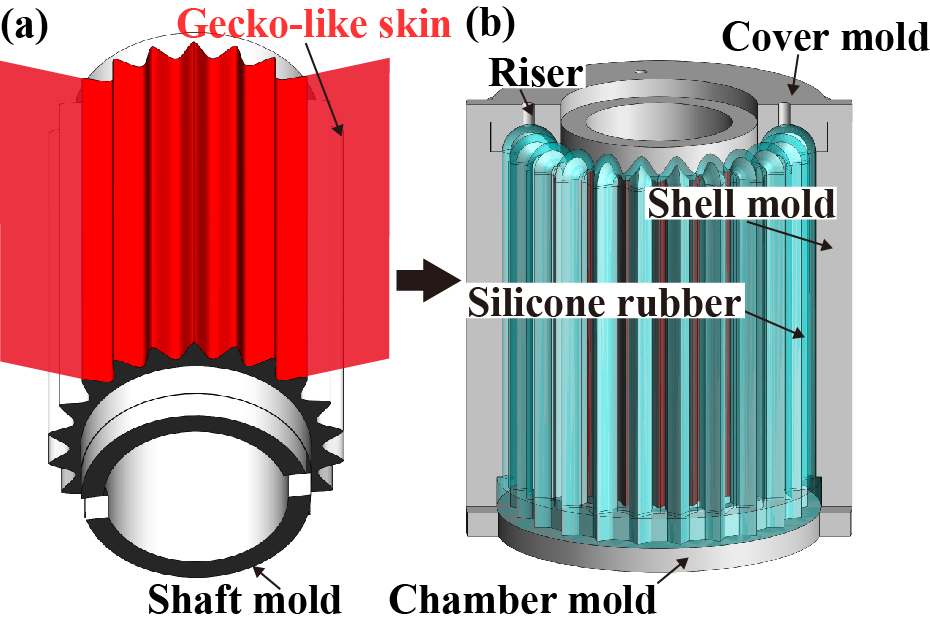}
	\caption{
		The fabrication of the gripper is based on a multi-part casting procedure. (a) Transfer printing is used to apply the patterned gecko-like skin to the surface of the shaft mold. (b) After assembling the chamber mold, shell mold, and shaft mold, uncured liquid silicone rubber is poured into the mold assembly, which is completed via a cover mold.
	}
	\label{fig:fabrication} 
\end{figure}

\begin{figure*}[t!]
	\centering
	\includegraphics[width=150mm]{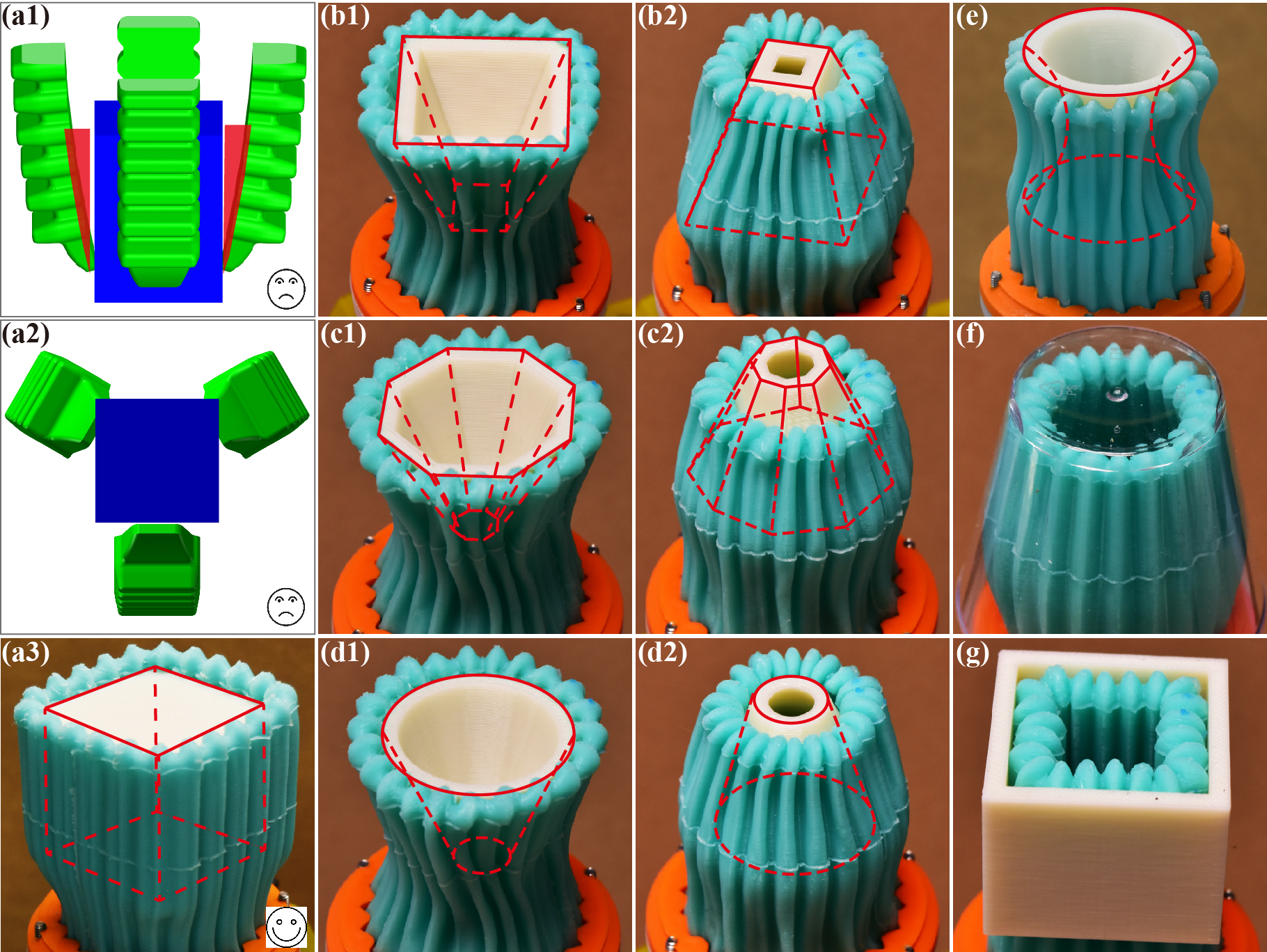}
	\caption{
		 The performance of our gripper is facilitated by its ability to passively conform to the shape of grasped objects.  It can be compared with grasping via fingered grippers, (a1) which often involve gaps between the fingers and object. (a2) Grasping via fingered grippers also demands that the object configuration and finger poses be accounted for, limiting the range of feasible grasp poses. (a3) In contrast, our gripper can readily conform to the shape of the object.  Using our   enveloping method, many object shapes can be  grasped, such as a rectangular pyramid (b1), a hexagonal pyramid (c1), or a cone (d1). The same objects can be grasped when inverted (b2), (c2) and (d2).  The gripper can also conform to  objects with concave shapes (e). By expanding into an orifice, the gripper can grasp objects via their interior, such as a cylinder (f) or cube  (g).
	}
	\label{fig:shape_adaption} 
\end{figure*}

\section{Materials and Methods}


\subsection{Design and Fabrication}
The design of the gripper is based on a cast silicone structure that enables the gripper to perform multi-functional grasping via a single fluidic port (Fig.~\ref{fig:design_and_ principal}). The gripper is composed of a main accordion structure which contains 22 parallel chambers connected to a single fluidic port.  Three fasteners  are used to seal, pressurize and vacuum the gripper (Fig.~\ref{fig:design_and_ principal}a2). To increase the load capacity of the gripper, we add a gecko-like skin layer to the inner surface(Fig.~\ref{fig:design_and_ principal}a3). The chambers form a cylindrical aperture. Under deflation, the side walls fold together and the aperture consequently shrinks to produce a smaller circumference (Fig.~\ref{fig:design_and_ principal}a1). Under inflation, the distance between the walls increases, causing the aperture to expand to a bigger cup, similar in shape to a lotus flower (Fig.~\ref{fig:design_and_ principal}a3). The motion of the gripper is illustrated in Supplementary Video S1. A cross-sectional view also illustrates the working principle (Fig.~\ref{fig:design_and_ principal}b1-b3). These inflation and deflation behaviors enable two modes of grasping, based on contracting around an object or inflating within an object aperture.  Application of positive pressure enables the gripper to ``swallow'' the object, while subsequent application of vacuum pressure causes it to envelop and conform to the object for lifting. This process is reversed in the interior grasping mode, for which the gripper is first deflated to insert into an aperture in the object, then expanded in order to grasp it.

The gripper is primarily molded from low viscosity platinum catalyzed silicone polymer (Mold Star 15, Smooth-On Inc., USA). The fabrication process is illustrated in Fig.~\ref{fig:fabrication}. After producing the molds, we fabricate the gecko-like skin via a  mold with a patterned texture \cite{suresh2019spatially}. We transfer print this skin onto the wall of the shaft mold (Fig.~\ref{fig:fabrication}a). Subsequently, we assemble the shaft mold, chamber mold and shell mold together and fill it with  uncured silicone rubber. After degassing in the vacuum container for about thirty minutes, we place a cover mold on top of the mold assembly and allow it to cure for eight hours. A small amount of excess silicone rubber is expelled through the riser (Fig.~\ref{fig:fabrication}b).

\begin{figure}[!ht]
	\centering
	\includegraphics[width=75mm]{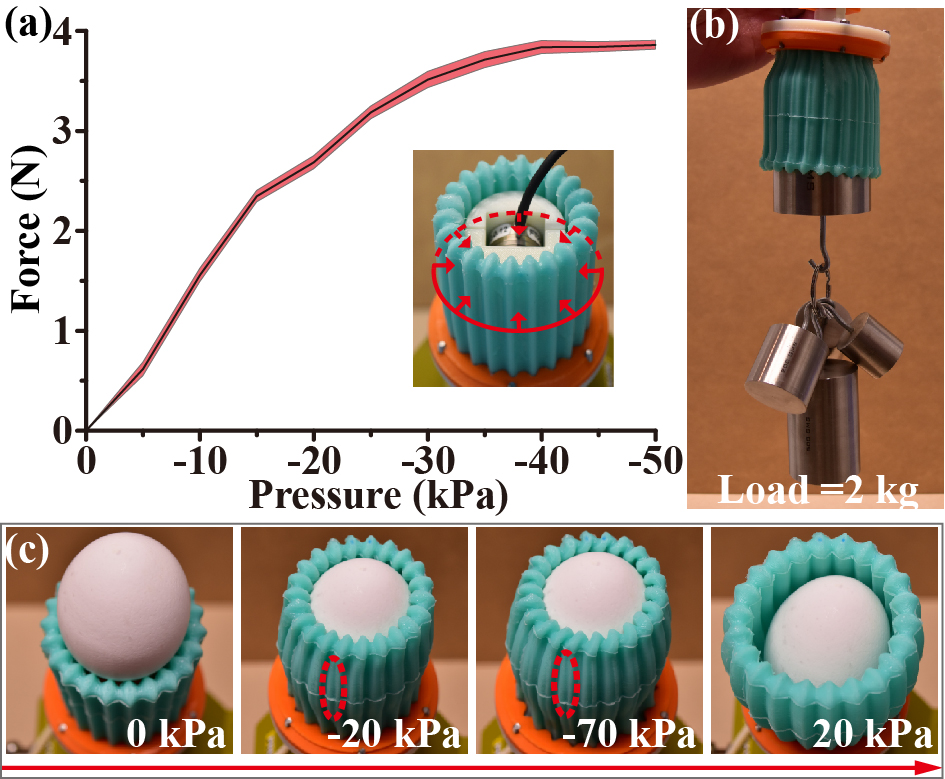}
	\caption{
		The gripper can not only produce substantial lifting forces, but can also safely handle fragile objects by virtue of a wall buckling process. (a) Squeezing forces are produced through the application of vacuum pressure. During object grasping, the forces increase little beyond a vacuum pressure level of -30 kPa, at which point the interior chambers buckle and collapse, ensuring safe grasping. The diameter of the object is 32 mm. (b) The gripper can lift a weight of 2 kg owning to the gecko-like adhesion. (c) The gripper does not crush the egg even under the application of  high vacuum pressures due to the buckling effect. The red dashed line illustrates the collapse of a chamber.
	}
	\label{fig:advantage} 
\end{figure}

\subsection{Passive Shape Conformation, Wall Buckling and Gecko-inspired Adhesion}
The ability of the gripper to passively conform to objects of various sizes, shapes and poses facilitates  grasping. Grasping with fingered grippers often produces gaps, due to the geometry and bending action of the fingers (Fig.~\ref{fig:shape_adaption}a1). For such grippers, the orientations of the fingers and objects must be accounted for in order to achieve stable grasping (Fig.~\ref{fig:shape_adaption}a2). In contrast, our gripper can readily conform to objects in various shapes, sizes, and poses by passively adapting to their shapes (Fig.~\ref{fig:shape_adaption}a3-g). When the gripper is actuated to grasp an  object, points on the gripper surface stay in place upon contacting the object, while other surface points on the gripper continue to move until a large grasp surface is produced.  When deflated, the gripper can adapt to  the  shape of a variety of objects; Fig.~\ref{fig:shape_adaption}b1 to d2 demonstrates adaptation to   rectangular and hexagonal pyramids or cones, even when these objects are placed upside down. The gripper can also grasp objects via concave surfaces (Fig.~\ref{fig:shape_adaption}c). In the expansion-based grasping mode, the gripper is inserted into an orifice and inflated to conform to the  apertures of objects, such as cups (Fig.~\ref{fig:shape_adaption}f) or hollow cubes (Fig.~\ref{fig:shape_adaption}g). It is challenging to achieve similar levels of shape adaptation via fingered grippers.

The safety of soft grippers is greatly facilitated by the softness of the materials. However, these materials often restrict load capacity. One way to address this is by dynamically controlling stiffness. However, increasing stiffness can negatively impact safety, causing a gripper to damage fragile objects, and may also degrade the ability of the gripper to conform to objects.  Our device can safely grasp brittle objects without excessive decrease in load capacity by virtue of the synergistic combination of wall buckling and gecko-like adhesion properties that are integrated in the design. When deflated, the gripper shrinks to squeeze the object. However, the squeezing force plateaus as pressure is decreased further, due to the collapse of the pneumatic chambers in the walls, as our results show (Fig.~\ref{fig:advantage}a). This wall buckling prevents the gripper from damaging fragile objects, such as eggs (Fig.~\ref{fig:advantage}c), even under the application of large negative pressures.  The  gecko-like skin that patterns the interior surface ensures that the load capacity of the gripper is nonetheless sufficient to lift objects many times heavier than the gripper itself, up to 2 kg in mass (Fig.~\ref{fig:advantage}b).  Such loads are often prohibitively large for silicone rubber grippers to lift. The safety and load capacity of the gripper are further illustrated in Supplementary Video S2. 
The same attribute that enables the safety and load capacity of the gripper ensures that the grasping force needs not be adjusted continuously.  In contrast, many soft grippers require pressure to be continuously controlled so that appropriate grasp forces may be applied.  Likewise, the gripper needs not be operated in a manner that varies with the shape of the grasped object, due to the conforming property of the grasping operation.  Together, these properties allow this gripper to be operated via a very simple fluidic input, making it amenable to many practical applications.

\subsection{Analytical Model}
\begin{figure*}[t!]
	\centering
	\includegraphics[width=150mm]{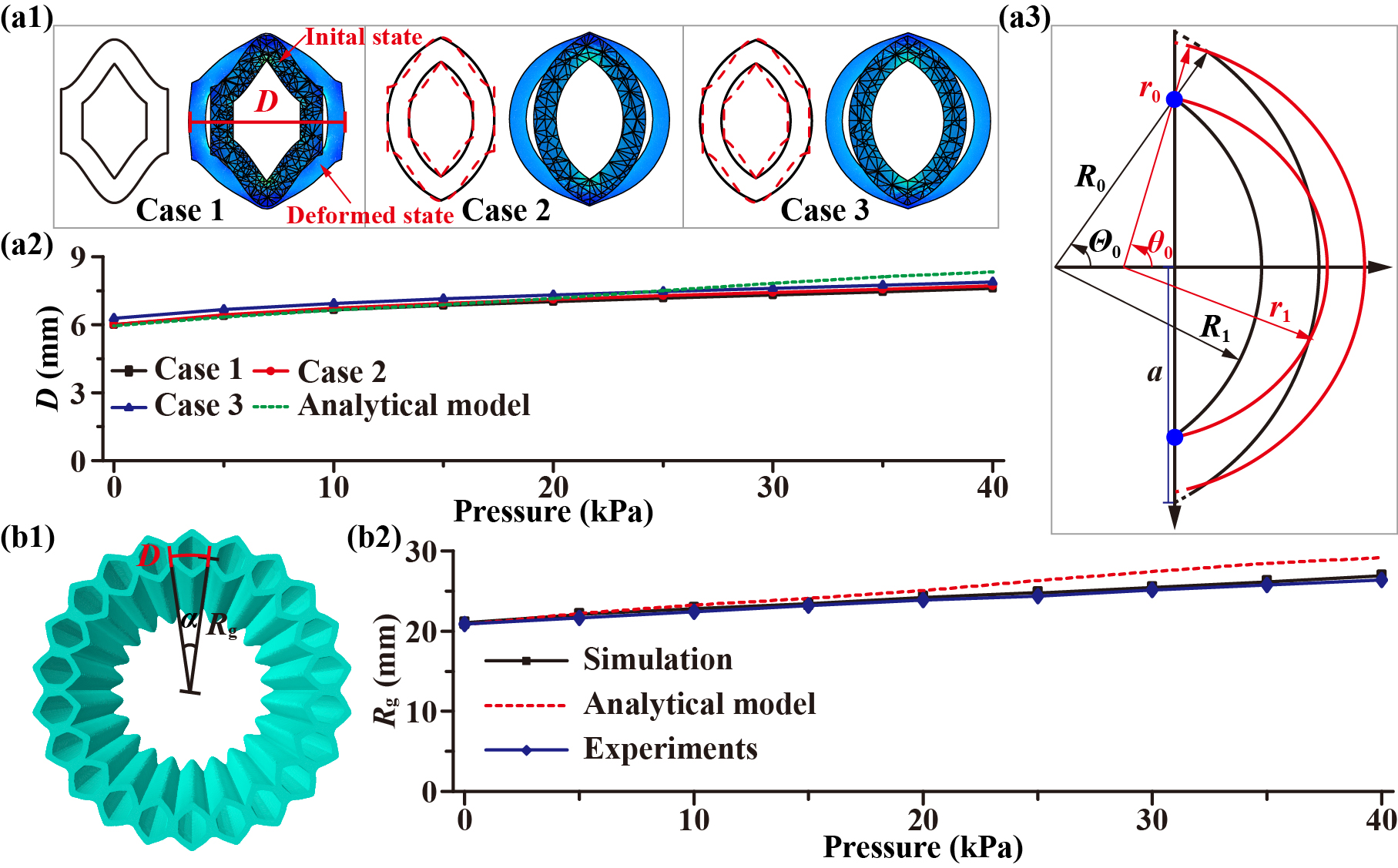}
	\caption{
		Predictions of the analytical model, numerical simulation, and measured behavior of the gripper as functions of fluid pressure. (a1) Matching single chamber geometry to the model, section views.   Case 1: True geometry. Case 2: Approximation via circular arcs tangent to the chamber. Case 3: Approximation via arcs constrained at the symmetric top and bottom corners of the chamber. $D$: Interior width. (a2) Values of the width $D$ for numerical models (cases 1, 2, 3) and the analytical model for pressures from 0 to 40 kPa.  (a3) Simplified geometry of the half wall of one chamber as used in the  analytical model. (b1) The geometric relationship of the deformed gripper.  $R_g$ is the gripper aperture radius.  (b2) Comparison of gripping aperture size vs.~pressure for the simulation, analytical model, and measurements. 
	}
	\label{fig:mathematical_model} 
\end{figure*}

The workspace of the gripper determines the smallest and largest objects it can grasp. The smallest object that can be grasped is determined by the minimum gripper aperture. This depends on the number and dimensions (principally thickness) of the chamber. The largest graspable object size is determined by the maximum gripper aperture, which occurs at a pressure for which the accordion structure becomes maximally unfolded. We developed an analytical model to predict the diameter of the gripper as a function of applied pressure. Because all  chambers have the same geometry, we first established a relationship between the pressure and the geometry of a single chamber (the red line in Fig.~\ref{fig:mathematical_model}a1), and extrapolate from this to determine the shape of the entire gripper aperture.

The geometry of the chamber (Case 1 in Fig.~\ref{fig:mathematical_model}a1) is complex, so we approximated the profile as circular arcs which captured the characteristic deformation. We considered two approximations as circular arcs (Fig.~\ref{fig:mathematical_model}a1).  One in which the arcs were tangent to the wall of the chamber at the midpoint (Case 2), and one in which the arcs circumscribed the geometry of the wall along the entire length (Case 3). 
From numerical finite element analysis  (FEA) of the true chamber geometry (Case 1) and the geometric approximations (Cases 2, 3), we determined that the error of approximation was smaller for Case 2 than for Case 3 (1.46\% vs.~3.78\%), so we used this approximation in developing our analytical model.  

For reasons of symmetry, we can focus on half of the chamber (Fig.~\ref{fig:mathematical_model}a3).   For moderate pressures, the walls undergo little stretching, so we ignore the strain along the length of the walls, and assume that the cross section area of the chamber section remains constant.  Our simulation revealed that the top and bottom edges of the inner surface (blue points in Fig.~\ref{fig:mathematical_model}a3) displaced vertically by only 0.4 mm when the pressure changed from 0 to 40 kPa, so we constrained these points to be fixed in space for modeling simplicity.  We refer to the uninflated inner and outer radii as $R_{1}$ and $R_{0}$ respectively, and the half central angle as $\bigtheta{0}$. After inflation, the inner radius, outer radius, and half central angle change to $r_{1}$, $r_{0}$, and $\theta_{0}$ separately. The principle strain in the $\theta$ direction is
\begin{equation}
\lambda_\theta=r\theta_0/(R\bigtheta{0}). 
\end{equation}
Because we assume that the material is incompressible, the principle strain in the $r$ direction is
\begin{equation}
\lambda_r=R\Theta_0/(r\theta_0). 
\end{equation}
Applying the Cauchy equilibrium equations, we obtain
\begin{equation}
d\sigma_{rr}/dr=(\sigma_{\theta\theta}-\sigma_{rr})/r, 
\end{equation}
where $\sigma_{rr}$ is the stress in the $r$ direction, and $\sigma_{\theta\theta}$ is the stress in the $\theta$ direction. Force balance in the $r$ direction implies that
\begin{equation}
P=\int_{r_1}^{r_0}(\sigma_{\theta\theta}-\sigma_{rr})/r\ dr, 
\end{equation}
where $P$ is the inflating pressure.

The relationship between stress and strain is determined by the material properties. Here we adopt an incompressible, neo-Hookean model for the elastomer (Mold Star 15) \cite{kidambi2018modular}. The strain energy density function for the material is
\begin{equation}
W=C_1(I_1-3), 
\end{equation}
where $C_1$ is a material constant with a value of 119 kPa \cite{kidambi2018modular}, and $I_1$ is the first invariant of the left Cauchy-Green deformation tensor
\begin{equation}
I_1=\lambda_1^2+\lambda_2^2+\lambda_3^2.
\end{equation}
The Cauchy stress difference for $\sigma_{\theta\theta}$ and $\sigma_{rr}$ is given by
\begin{equation}
\sigma_{\theta\theta}-\sigma_{rr}=\lambda_{\theta}(\partial W/\partial\lambda_{\theta})-\lambda_{r}(\partial W/\partial\lambda_{r}),
\end{equation}
where $\partial W/\partial\lambda_{\theta}=2C_1\lambda_{\theta}$, and $\partial W/\partial\lambda_{r}=2C_1\lambda_{r}$. Substituting these two equations into equation 7, we  obtain
\begin{equation}
\sigma_{\theta\theta}-\sigma_{rr}=2C_1\lambda_{\theta}^2-2C_1\lambda_{r}^2.
\end{equation}
The deformation of the chamber is described by
\begin{equation}
(R^2-R_1^2)\bigtheta{0}=(r^2-r_1^2)\theta_{0}.
\end{equation}
Substituting equation 1, 2, 8 and 9 into equation 4, we obtain
\begin{equation}
\begin{split}
P=2C_1{\frac{\theta_{0}}{\bigtheta{0}}}\ln\frac{R_0}{R_1}+&C1{\frac{\bigtheta{0}}{\theta_{0}^2}}(R_1^2\bigtheta{0}-r_1^2\theta_{0})(\frac{1}{r_0^2}-\frac{1}{r_1^2})\\
&-C_1\frac{\bigtheta{0}}{\theta_{0}}\ln\frac{r_0}{r_1}.
\end{split}
\end{equation}
The edges (blue points in Fig.~\ref{fig:mathematical_model}a3) are fixed, thus
\begin{equation}
r_1\sin\theta_{0}=a,
\end{equation}
where $a$ is  half of the initial distance between the two end points.
The cross sectional area may be assumed to be constant:
\begin{equation}
(R_0^2-R_1^2)\bigtheta{0}=(r_0^2-r_1^2)\theta_{0}.
\end{equation}
Solving the nonlinear set of equations 10, 11 and 12, yields expressions for $r_0$, $r_1$, and $\theta_{0}$. We solved these numerically.  In order to find solutions matching our configuration, we restricted their ranges to values near our design, i.e. 
\begin{equation}
r_0\in[4.56, 5], \ \
r_1\in[3, 3.8], \ \ 
\theta_0\in[57.6^\circ, 80^\circ].
\end{equation}
The expression for the wall distance, \textit{D}, is then given by:
\begin{equation}
D=2(r_0-r_1\cos\theta_0).
\end{equation}
As Fig.~\ref{fig:mathematical_model}a2 shows, the resulting analytical model exhibits excellent agreement with the FEA simulation for pressures below 20 kPa. The maximum error is approximately 0.04 mm. Larger errors occur at higher pressures. We speculate this may be caused by the fixed edge in our analytical model, since simulations showed that this edge moved by a small amount when pressure was higher.

The gripper has twenty two identical chambers. When the gripper is inflated, the arc length of one chamber \textit{D} and expanding radius \textit{R}$\rm_g$ will increase while the central angle $\alpha$ remains constant (Fig.~\ref{fig:mathematical_model}b1). From this, we can obtain the relationship between \textit{D} and \textit{R}$\rm_g$. Here, \textit{D} is adopted from Fig.~\ref{fig:mathematical_model}a1 to approximate the length of the arc, and $\alpha=$360$^\circ$/22. We have
\begin{equation}
R_{\rm g}=D/\alpha.
\end{equation}

To evaluate the analytical model, we compared the predicted dependence of \textit{R}$\rm_g$ on pressure with the numerical FEA simulation and with  laboratory experiments. For the FEA simulation, we inflated the gripper to different pressures and determined the coordinates of the center points of the chambers, then used these coordinates to fit the corresponding circles and used this to calculate the values of \textit{R}$\rm_g$. For the experiments, we inflated the gripper to different pressures and measured the deformed shape. After digitizing the center points of the chambers in software, we calculated values of \textit{R}$\rm_g$. Results from Fig.~\ref{fig:mathematical_model}b2 show that the maximum error for the simulation in comparison with the experiments is 0.49 mm, which shows that the simulation agrees  with the experiments. For the analytical model, the error is less than 1.2 mm when the pressure is below  20 kPa. However, as in the case analyzed above, the error increases with  increasing  pressure. Two further factors may explain the errors. One is that  errors were incurred when calculating \textit{D}. The other is that we neglected  interactions between the neighboring chambers. Nonetheless, the results show good qualitative agreement across a wide range of fluidic pressures.

\section{Results and Discussion} 
\subsection{Gripping Force via Contraction-based Grasping}

\begin{figure*}[t!]
	\centering
	\includegraphics[width=150mm]{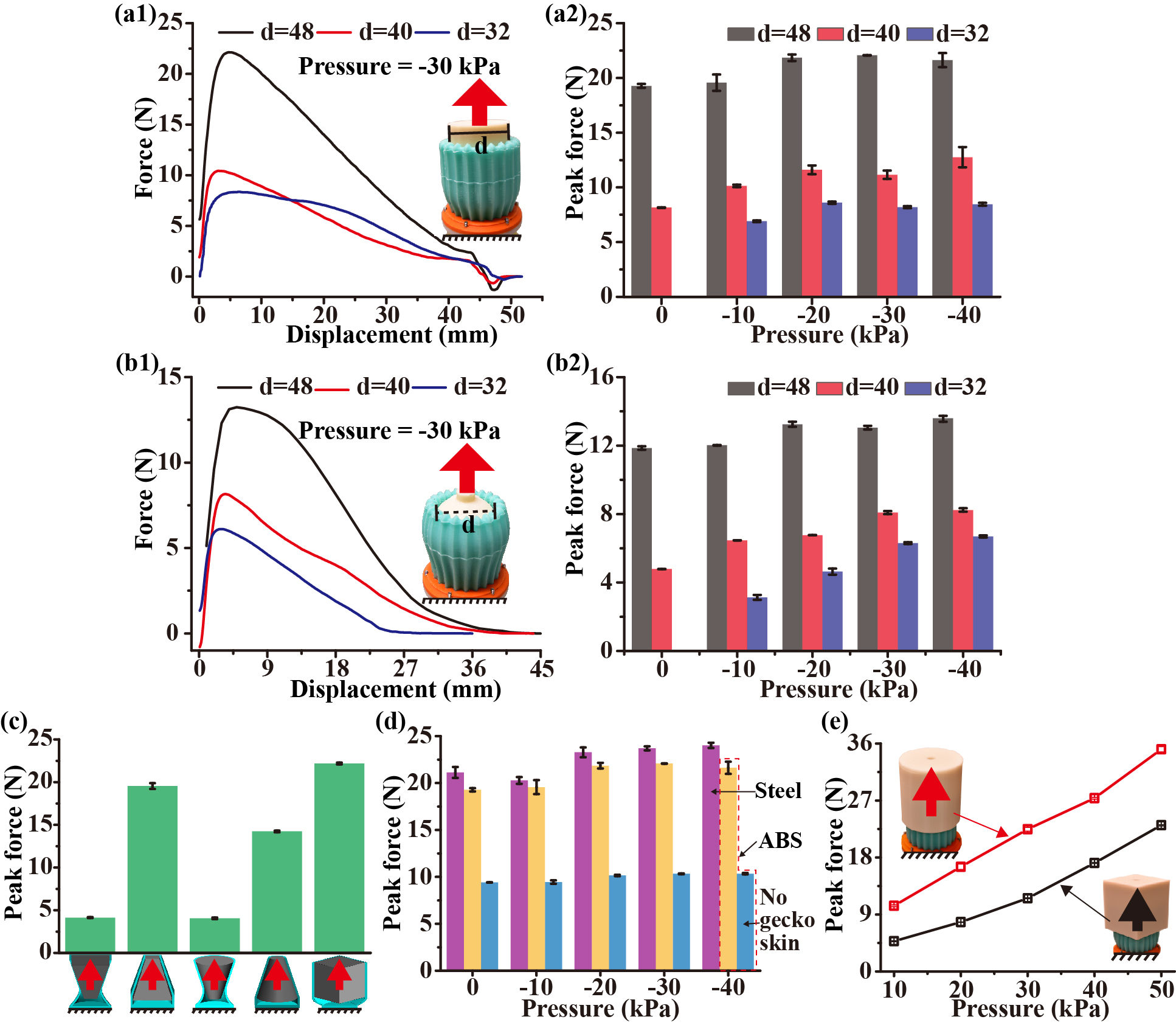}
	\caption{
		Holding force results for the gripper. (a1) Measured  force as the cylinders of different diameters are gradually pulled up out of the gripper. The air pressure was constant, -30 kPa, for all the objects. (a2) The peak forces for the three cylinders under different air pressures. (b1) Measured force as the spheres of different diameters are gradually pulled upward out of the gripper. The air pressure was -30 kPa for all the objects. (b2) The peak forces for the three spheres with varying fluid pressure. (c) The peak forces for objects of different shapes and poses at fixed pressure -30 kPa. All the objects have the same inscribed circle. (d) The comparison of the peak forces when the gripper grasped the 48mm cylinders with different surface roughness (steel and 3D printed ABS), and when the grippers with gecko-like skin and without it grasp the same ABS cylinder. The diameter of the cylinders was 48 mm. (e) The peak forces of the cylinder and the square in the expansion-based
		 grasping mode.
	}
	\label{fig:holding_force} 
\end{figure*}

To investigate the factors affecting the performance of the gripper, we conducted experiments to assess the dependence of gripping force on the size and shape of objects, the actuating pressure, and the interactions between gecko-like adhesion and surface roughness. Fig.~\ref{fig:holding_force}a1 shows that the lifting force increases quickly then decreases slowly as a cylinder detaches from the gripper. For all sizes of cylinder, the  pressures required for lifting are identical (-30 kPa). As the cylinders were lifted, the features on the gecko skin deformed to ensure that the load was evenly shared across a large contact area, increasing the maximum friction force. When the diameter increases from 32 mm to 48 mm, the peak force also increases. This is because the contact area increases with  increasing  diameter. Fig.~\ref{fig:holding_force}a2 illustrates the peak force as a function of vacuum pressure for all the three cylinders. When the gripper is deflated from 0 to -20 kPa, the peak forces increased similarly for all the cylinders. Subsequently, the peak forces change little as the vacuum is augmented. This is a result of the wall buckling of the chambers. It should be noted that the data for the 32 mm diameter under 0 kPa is missing because the gripper didn't contact the object in this case.

When the gripper grasped spheres of different diameters, the results we obtained are similar to those that we observed for the cylinders (Fig.~\ref{fig:holding_force}b1).  The peak forces stop increasing when the pressure reaches a threshold  value because of the wall buckling effect (Fig.~\ref{fig:holding_force}b2). For the 48 mm sphere, this value is -20 kPa, while for the 40 mm and 32 mm spheres, the value is -30 kPa. Comparing the results shown in Fig.~\ref{fig:holding_force}b and Fig.~\ref{fig:holding_force}a, the forces produced when gripping the spheres are smaller than those for similarly sized cylinders at the same  pressure. Thus, apart from the size, the gripping force of the gripper also depends on the shape of the grasped objects, as expected. This is also confirmed in the results shown in Fig.~\ref{fig:holding_force}c. At the same driving pressure (-30 kPa), the peak forces are different for each of the rectangular pyramid, the cone and the cube because the contact areas are different. When the gripper graspes the  rectangular pyramid and cone, the peak forces also change significantly when the objects were grasped  upside down. For the rectangular pyramid with the larger area facing down (the second object  from the left in Fig.~\ref{fig:holding_force}c), the peak force is nearly four times of that when the object is upside down (see leftmost object in Fig.~\ref{fig:holding_force}c).

\subsection{Effects of Gecko-like Adhesion and Expansion-based Grasping}  


To investigate the effects of the gecko-like skin that was applied to the interior of the gripper, we tested the gripping forces of two grippers (one with gecko-like skin while the other without it) as they grasped the same cylinder with a diameter of 48 mm. The object  was 3D printed using ABS plastic. We also measured the gripping forces when the gripper with gecko-like skin grasped two 48 mm diameter cylinders with different surface roughness. In addition to  the ABS plastic cylinder, we also prepared a steel cylinder whose surface was smoother. The results are shown in Fig.~\ref{fig:holding_force}d. When grasping the  ABS cylinder using specified pressure, the peak forces of the gripper with gecko-like skin are nearly twice those of the gripper that lacked the skin. Thus, the gecko-like surface increases the gripping force. When a smooth steel object was grasped, the peak forces of the gripper with gecko-like skin are larger than those produced when grasping the rougher object made of ABS plastic. However, the differences are small, suggesting that  such a gecko-like skin is effective in many dry surface conditions. It is worth noting that when the pressure is 0 kPa, large peak forces are produced provided the diameters of the objects are larger than the uninflated interior diameter of the gripper (Fig.~\ref{fig:holding_force}a2, b2 and d). For example, in Fig.~\ref{fig:holding_force}d, the peak forces of the gripper with gecko-like skin are nearly 20 N. This occurs because when grasping a object with a diameter bigger than the nominal uninflated diameter of the gripper, the gripper can be  stretched after contacting the target, which enables it  to produce friction forces sufficient to lift  the object. Consequently, the gripper can manipulate some large objects without the use of fluid energy. We also measured the gripping forces in the expanding grasping mode. As the results in Fig.~\ref{fig:holding_force}e show, the peak forces increase as the gripper is  inflated to higher pressure. This occurs because a larger air pressure  generates a larger normal force against the object. There is no gecko-like skin on the outside surface of the gripper, although such a skin could also be applied to the exterior surface. Comparing the results obtained for the cylinder with those for the cube,  the peak force varied with the object shape. This occurred because the contact areas depend on the object shape.

\begin{figure*}[t!]
	\centering
	\includegraphics[width=150mm]{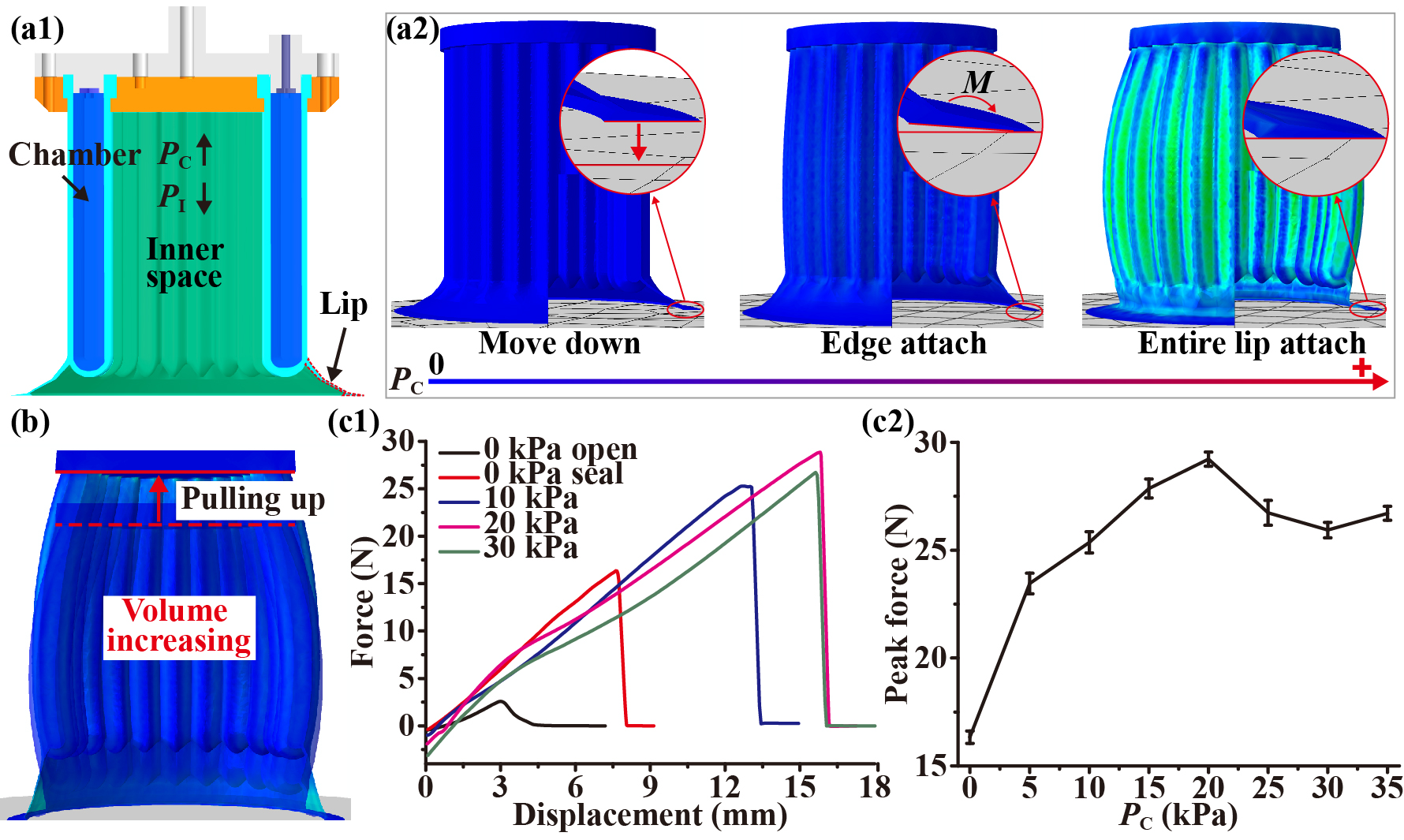}
	\caption{
		The expansion-driven suction mode of grasping. (a1) Operating principle. When the gripper is inflated, the pressure, \textit{P}$\rm_C$, of the chambers increases causing the lip to form a seal with the inner space. As the volume of the inner space increases, the pressure, \textit{P}$\rm_I$, inside that  space decreases. (a2) The FEA simulation shows how the seal is generated via inflation. (b) Because the gripper is soft, the volume of the inner space increases further when the gripper is raised, enhancing the suction force. (c1) Force vs.~pressure, \textit{P}$\rm_C$, as the gripper is raised. (c2) The peak forces observed at different pressures, \textit{P}$\rm_C$.
	}
	\label{fig:suction_force} 
\end{figure*}

\subsection{Gripping Flat Objects via Expansion-Driven Suction} 

We augmented the gripper with a lip at the opening that enables the gripper to produce suction sufficient for lifting flat objects.  Suction is generated through volume changes in the region enclosed by the inner surface of the gripper (Fig.~\ref{fig:suction_force}a1 and a2).  As the pressure, \textit{P}$\rm_C$, of the chambers is increased, the lip bends in contact with a flat surface, forming a seal with the interior space. The wedge shape of the lip ensures that the exterior part contacts the surface first, avoiding wrinkling or other local deformations that would inhibit seal formation.  As Fig.~\ref{fig:suction_force}a2 shows, when the chambers of the gripper are inflated, a moment is generated that produces a downward tilt in the outer rim of the lip.  Further inflation reinforces the seal with the flat surface.  As the chamber pressure continues to increase, the gripper inflates in a manner that causes an increase in the enclosed volume, thereby decreasing the pressure in the enclosed region, generating suction.  Because the gripper is soft, the volume of the inner space increases further when the gripper is raised, enhancing the suction force (Fig.~\ref{fig:suction_force}b). We measured the suction force when the gripper chambers were inflated to different pressures. The suction force increases as the gripper is raised, before suddenly decreasing to zero when the maximum force it can exert on the surface is reached.  If no seal is formed, the maximum lifting force that can be generated  is just 2 N.  When a seal is formed, this rises to 15 N (Fig.~\ref{fig:suction_force}b), demonstrating that pulling the gripper upward after sealing generates suction through the expansion that is produced in the interior volume via lifting.  When the chambers are pressurized, the increased suction that is produced causes the maximum lifting force to rise  to nearly 30 N at 20 kPa.

This mode of operation may be compared to conventional vacuum-driven suction cups, which have been widely used.  Such suction cups typically require a dedicated fluid port, and often demand continuous airflow during  operation.  In contrast, our method requires no bulk fluid flow, since suction is generated via the low-volume inflation of the gripper, using the same fluid port we use for the other operating modes of the gripper (i.e., the contraction and expansion modes).  Thus, this mode of operation does not add to the complexity of the system.  The presence of the lip does affect the profile of the gripper, which may have some effect on the other grasping modes, depending on the objects involved.  However, this can be addressed, if needed, by optimizing the lip geometry (for example, using a lower profile or thinner lip), or through the use of a modular lip that may be detached.

\begin{figure*}[t!]
	\centering
	\includegraphics[width=150mm]{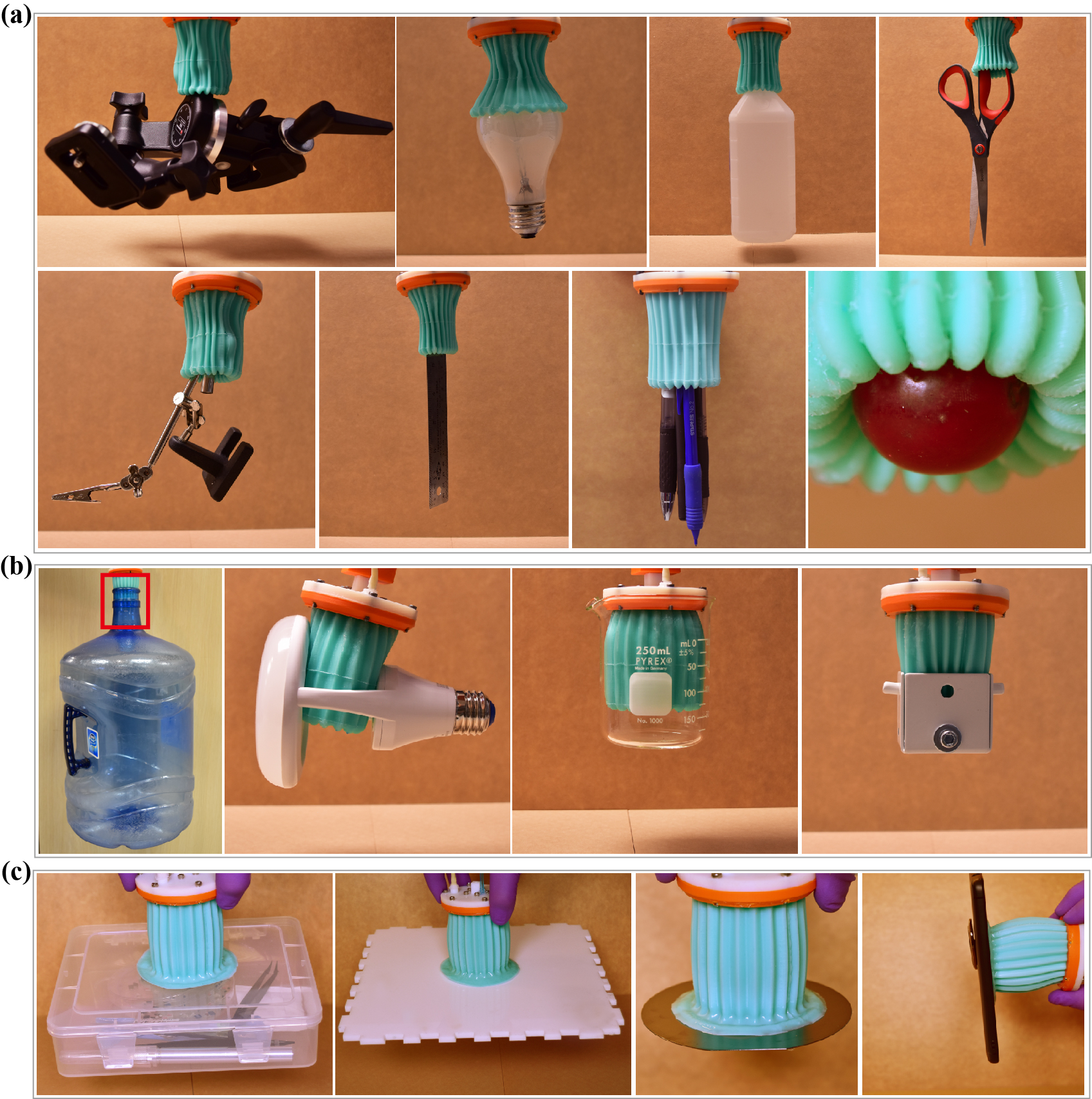}
	\caption{
		Examples of the performance of the gripper in different grasping modes. (a) Contraction-based grasping of a large camera clamp (1 kg), a light bulb, a bottle of isopropyl alcohol (400g), scissors, a soldering clamp, a ruler, a bundle of pens, or a grape. (b) Expansion-based grasping of a water dispenser barrel, a light bulb, a beaker, or a hollow cube. (c) Expansion-driven suction based grasping of a box containing several objects, an acrylic plate, a silicon wafer, and a mobile phone.
	}
	\label{fig:grasping_object} 
\end{figure*}

\subsection{Multimodal Grasping} 

Through compliant shape-adaptation and gecko-like adhesion, the gripper can grasp multiple objects when driven via a simple inflating-deflating control from a single port. Similar grasping performance can be achieved irrespective of variations in the detailed geometry or pose of the objects, 
and the driving pressure. When grasping via expansion-driven suction, the gripper can also attach to and lift flat objects without continuous vacuuming. To confirm this, we conducted experiments on grasping a large variety of objects using different strategies. When grasping objects via contraction, the gripper was inflated to a maximum pressure of 40 kPa, and subsequently deflated to -40 kPa to enclose the objects. Results from Fig.~\ref{fig:grasping_object}a show that the gripper can grasp a large camera clamp (1 kg), a bottle, scissors, a soldering clamp, a ruler, a bundle of pens, or a grape using the same simple behavior. Using the gecko-like skin, it can grasp a light bulb via the convex  upper surface, without fully enclosing middle of the object. Grasping can be achieved provided the object, or a protuberant part of the object, is within the workspace of the gripper and the weight does not exceed the force capacity (which depends on the object geometry, see above). In contrast, when a fingered gripper is used to grasp objects such as scissors, soldering clamp, or ruler, computation is required in order to determine feasible grasping locations at which the gripper can achieve force or form closure.  For a wide range of objects, our gripper is able to grasp without such calculations, greatly simplifying the process.

Via the expansion-based grasping mode, the gripper is inserted within an inner surface, handle,  or aperture and inflated in order to produce grasp forces. As is shown in Fig.~\ref{fig:grasping_object}b, using this mode, the gripper can grasp a water dispenser barrel, a light bulb, a beaker or a hollow cube via different openings in each. This only requires that a suitable cavity exists in which the gripper can be inserted. In Supplementary Video S3, we show how the gripper is able to perform these operations.  We also show how,  for some objects, multiple successful grasp strategies exist, through which the gripper conforms to different protuberances, achieving grasping via expansion or contraction. Fig.~\ref{fig:grasping_object}c shows that inflating the gripper enables it to use suction to attach to a variety of objects with flat surfaces: a box, a plate, a silicon wafer, or a mobile phone. The process through which each of these objects was grasped is shown in Supplementary Video S4.

Although the gripper can grasp many kinds of objects with a simple control, using fluid from a single port, there are several conditions in which its performance can be improved through further research. For example, the gripper cannot easily grasp a small object, such as a pen, lying on a table or a rough surface that cannot be sealed. To achieve successful grasping, the size of the objects or the protuberance must be within a feasible  range. For the prototype used in our experiments, the typical range of object sizes that can be grasped via contraction is 1 cm to 5 cm, although larger curved objects can also be grasped. The feasible object dimensions are determined by the expanding and folding ratio. The weight of the object to be grasped is also a determining factor, as illustrated in the experiments described in the preceding sections. For an object of a particular size, shape, and pose, a maximum load capacity must be respected.

\section{Conclusion}
In summary, this paper presents a new fingerless soft gripper capable of efficiently generating multiple grasping modes.  It is based on a soft accordion structure containing coupled, parallel fluidic channels. This structure allows the gripper to passively adapt its shape to conform to grasped objects.  It is controlled via pressure supplied from a single fluidic port. Inflation opens the gripper orifice for enveloping an object, while deflation produces grasping forces.  The interior is patterned with a gecko-like skin that increases friction, enabling the gripper to lift objects weighing up to 20 N without continuously applied power.  Such lifting forces are larger than those that can be produced by many fingered grippers.  Our design ensures that fragile objects, such as eggs, can be safely handled by virtue of a wall buckling mechanism. The gripper also admits a mode of grasping in which it may be inflated within an opening or orifice.  This enables it to grasp objects with handles or openings. We also show how the design of an integrated lip allows the gripper to form a seal, and, upon inflating, to generate suction sufficient to lift many flat objects.  This simple design thus integrates features that ensure it can conform to many objects via different features or flat surfaces. 

The parsimonious combination of features, and the grasping modes they enable, allows this gripper design to solve many grasping tasks via a single fluidic input.  This design is amenable for use in a wide range of tasks.  Such devices could improve the ability of robotic systems to meet application needs in areas of great economic and societal importance.  Some potential application domains include food processing, logistical sorting, including  pick-and-place sorting of heterogeneous objects on an assembly line.  There are  many opportunities for further extending the ideas presented here, in order to expand the workspace and capacity of the gripper. Such improvements could further expand the range of applications of such devices.


%

%

\section*{Acknowledgment}

This project is supported by China Scholarship Council fund to Y. H. (No. 201806020091), the US National Science Foundation under awards NSF-1628831, NSF-1623459, NSF-1751348 to Y.V. 

\ifCLASSOPTIONcaptionsoff
  \newpage
\fi



%
\bibliographystyle{IEEEtran}

\bibliography{IEEEabrv,refs}

\begin{thebibliography}{10}
\providecommand{\url}[1]{#1}
\csname url@samestyle\endcsname
\providecommand{\newblock}{\relax}
\providecommand{\bibinfo}[2]{#2}
\providecommand{\BIBentrySTDinterwordspacing}{\spaceskip=0pt\relax}
\providecommand{\BIBentryALTinterwordstretchfactor}{4}
\providecommand{\BIBentryALTinterwordspacing}{\spaceskip=\fontdimen2\font plus
\BIBentryALTinterwordstretchfactor\fontdimen3\font minus
  \fontdimen4\font\relax}
\providecommand{\BIBforeignlanguage}[2]{{%
\expandafter\ifx\csname l@#1\endcsname\relax
\typeout{** WARNING: IEEEtran.bst: No hyphenation pattern has been}%
\typeout{** loaded for the language `#1'. Using the pattern for}%
\typeout{** the default language instead.}%
\else
\language=\csname l@#1\endcsname
\fi
#2}}
\providecommand{\BIBdecl}{\relax}
\BIBdecl

\bibitem{melchiorri2008robot}
C.~Melchiorri and M.~Kaneko, ``Robot hands,'' \emph{Springer Handbook of
  Robotics}, pp. 345--360, 2008.

\bibitem{paolini2014data}
R.~Paolini, A.~Rodriguez, S.~S. Srinivasa, and M.~T. Mason, ``A data-driven
  statistical framework for post-grasp manipulation,'' \emph{Int. J. Robotics
  Res.}, vol.~33, no.~4, pp. 600--615, 2014.

\bibitem{pollard2004closure}
N.~S. Pollard, ``Closure and quality equivalence for efficient synthesis of
  grasps from examples,'' \emph{Int. J. Robotics Res.}, vol.~23, no.~6, pp.
  595--613, 2004.

\bibitem{ficuciello2019vision}
F.~Ficuciello, A.~Migliozzi, G.~Laudante, P.~Falco, and B.~Siciliano,
  ``Vision-based grasp learning of an anthropomorphic hand-arm system in a
  synergy-based control framework,'' \emph{Sci. Robot.}, vol.~4, p. eaao4900,
  2019.

\bibitem{shintake2018soft}
J.~Shintake, V.~Cacucciolo, D.~Floreano, and H.~Shea, ``Soft robotic
  grippers,'' \emph{Adv. Mater.}, vol.~30, no.~29, p. 1707035, 2018.

\bibitem{shintake2016versatile}
J.~Shintake, S.~Rosset, B.~Schubert, D.~Floreano, and H.~Shea, ``Versatile soft
  grippers with intrinsic electroadhesion based on multifunctional polymer
  actuators,'' \emph{Adv. Mater.}, vol.~28, no.~2, pp. 231--238, 2016.

\bibitem{lau2017dielectric}
G.-K. Lau, K.-R. Heng, A.~S. Ahmed, and M.~Shrestha, ``Dielectric elastomer
  fingers for versatile grasping and nimble pinching,'' \emph{Appl. Phys.
  Lett.}, vol. 110, no.~18, p. 182906, 2017.

\bibitem{shian2015dielectric}
S.~Shian, K.~Bertoldi, and D.~R. Clarke, ``Dielectric elastomer based
  “grippers” for soft robotics,'' \emph{Adv. Mater.}, vol.~27, no.~43, pp.
  6814--6819, 2015.

\bibitem{do2018miniature}
T.~N. Do, H.~Phan, T.-Q. Nguyen, and Y.~Visell, ``Miniature soft
  electromagnetic actuators for robotic applications,'' \emph{Adv. Funct.
  Mater.}, vol.~28, no.~18, p. 1800244, 2018.

\bibitem{diller2014three}
E.~Diller and M.~Sitti, ``Three-dimensional programmable assembly by untethered
  magnetic robotic micro-grippers,'' \emph{Adv. Funct. Mater.}, vol.~24,
  no.~28, pp. 4397--4404, 2014.

\bibitem{hubbard2019shrink}
A.~M. Hubbard, E.~Luong, A.~Ratanaphruks, R.~W. Mailen, J.~Genzer, and M.~D.
  Dickey, ``Shrink films get a grip,'' \emph{ACS Appl. Polym. Mater.}, vol.~1,
  no.~5, pp. 1088--1095, 2019.

\bibitem{breger2015self}
J.~C. Breger, C.~Yoon, R.~Xiao, H.~R. Kwag, M.~O. Wang, J.~P. Fisher, T.~D.
  Nguyen, and D.~H. Gracias, ``Self-folding thermo-magnetically responsive soft
  microgrippers,'' \emph{ACS Appl. Mater. Interfaces}, vol.~7, no.~5, pp.
  3398--3405, 2015.

\bibitem{pilz2019untethered}
M.~Pilz~da Cunha, Y.~Foelen, R.~J. van Raak, J.~N. Murphy, T.~A. Engels, M.~G.
  Debije, and A.~P. Schenning, ``An untethered magnetic-and light-responsive
  rotary gripper: Shedding light on photoresponsive liquid crystal actuators,''
  \emph{Adv. Opt. Mater.}, vol.~7, no.~7, p. 1801643, 2019.

\bibitem{abdullah2018self}
A.~M. Abdullah, X.~Li, P.~V. Braun, J.~A. Rogers, and K.~J. Hsia, ``Self-folded
  gripper-like architectures from stimuli-responsive bilayers,'' \emph{Adv.
  Mater.}, vol.~30, no.~31, p. 1801669, 2018.

\bibitem{zheng2018programmed}
S.~Y. Zheng, Y.~Shen, F.~Zhu, J.~Yin, J.~Qian, J.~Fu, Z.~L. Wu, and Q.~Zheng,
  ``Programmed deformations of 3d-printed tough physical hydrogels with high
  response speed and large output force,'' \emph{Adv. Funct. Mater.}, vol.~28,
  no.~37, p. 1803366, 2018.

\bibitem{mosadegh2014pneumatic}
B.~Mosadegh, P.~Polygerinos, C.~Keplinger, S.~Wennstedt, R.~F. Shepherd,
  U.~Gupta, J.~Shim, K.~Bertoldi, C.~J. Walsh, and G.~M. Whitesides,
  ``Pneumatic networks for soft robotics that actuate rapidly,'' \emph{Adv.
  Funct. Mater.}, vol.~24, no.~15, pp. 2163--2170, 2014.

\bibitem{uppalapati2018towards}
N.~K. Uppalapati and G.~Krishnan, ``Towards pneumatic spiral grippers: Modeling
  and design considerations,'' \emph{Soft Robot.}, vol.~5, no.~6, pp. 695--709,
  2018.

\bibitem{tawk2018bioinspired}
C.~Tawk, M.~in~het Panhuis, G.~M. Spinks, and G.~Alici, ``Bioinspired 3d
  printable soft vacuum actuators for locomotion robots, grippers and
  artificial muscles,'' \emph{Soft Robot.}, vol.~5, no.~6, pp. 685--694, 2018.

\bibitem{hao2018soft}
Y.~Hao, Z.~Gong, Z.~Xie, S.~Guan, X.~Yang, T.~Wang, and L.~Wen, ``A soft bionic
  gripper with variable effective length,'' \emph{J. Bionic Eng.}, vol.~15,
  no.~2, pp. 220--235, 2018.

\bibitem{galloway2016soft}
K.~C. Galloway, K.~P. Becker, B.~Phillips, J.~Kirby, S.~Licht, D.~Tchernov,
  R.~J. Wood, and D.~F. Gruber, ``Soft robotic grippers for biological sampling
  on deep reefs,'' \emph{Soft Robot.}, vol.~3, no.~1, pp. 23--33, 2016.

\bibitem{yuk2017hydraulic}
H.~Yuk, S.~Lin, C.~Ma, M.~Takaffoli, N.~X. Fang, and X.~Zhao, ``Hydraulic
  hydrogel actuators and robots optically and sonically camouflaged in water,''
  \emph{Nat. Commun.}, vol.~8, p. 14230, 2017.

\bibitem{hao2018eutectic}
Y.~Hao, T.~Wang, Z.~Xie, W.~Sun, Z.~Liu, X.~Fang, M.~Yang, and L.~Wen, ``A
  eutectic-alloy-infused soft actuator with sensing, tunable degrees of
  freedom, and stiffness properties,'' \emph{J. Micromech. Microeng}, vol.~28,
  no.~2, p. 024004, 2018.

\bibitem{zhou2018bcl}
J.~Zhou, J.~Yi, X.~Chen, Z.~Liu, and Z.~Wang, ``Bcl-13: A 13-dof soft robotic
  hand for dexterous grasping and in-hand manipulation,'' \emph{IEEE Robot.
  Autom. Lett.}, vol.~3, no.~4, pp. 3379--3386, 2018.

\bibitem{yang2017bioinspired}
Y.~Yang, Y.~Chen, Y.~Li, M.~Z. Chen, and Y.~Wei, ``Bioinspired robotic fingers
  based on pneumatic actuator and 3d printing of smart material,'' \emph{Soft
  Robot.}, vol.~4, no.~2, pp. 147--162, 2017.

\bibitem{wei2016novel}
Y.~Wei, Y.~Chen, T.~Ren, Q.~Chen, C.~Yan, Y.~Yang, and Y.~Li, ``A novel,
  variable stiffness robotic gripper based on integrated soft actuating and
  particle jamming,'' \emph{Soft Robot.}, vol.~3, no.~3, pp. 134--143, 2016.

\bibitem{jiang2019chain}
Y.~Jiang, D.~Chen, C.~Liu, and J.~Li, ``Chain-like granular jamming: A novel
  stiffness-programmable mechanism for soft robotics,'' \emph{Soft Robot.},
  vol.~6, no.~1, pp. 118--132, 2019.

\bibitem{yap2016high}
H.~K. Yap, H.~Y. Ng, and C.-H. Yeow, ``High-force soft printable pneumatics for
  soft robotic applications,'' \emph{Soft Robot.}, vol.~3, no.~3, pp. 144--158,
  2016.

\bibitem{zhu2019fully}
M.~Zhu, Y.~Mori, T.~Wakayama, A.~Wada, and S.~Kawamura, ``A fully
  multi-material three-dimensional printed soft gripper with variable stiffness
  for robust grasping,'' \emph{Soft Robot.}, 2019.

\bibitem{zhang2019fast}
Y.-F. Zhang, N.~Zhang, H.~Hingorani, N.~Ding, D.~Wang, C.~Yuan, B.~Zhang,
  G.~Gu, and Q.~Ge, ``Fast-response, stiffness-tunable soft actuator by hybrid
  multimaterial 3d printing,'' \emph{Adv. Funct. Mater.}, vol.~29, no.~15, p.
  1806698, 2019.

\bibitem{brown2010universal}
E.~Brown, N.~Rodenberg, J.~Amend, A.~Mozeika, E.~Steltz, M.~R. Zakin,
  H.~Lipson, and H.~M. Jaeger, ``Universal robotic gripper based on the jamming
  of granular material,'' \emph{Proc. Natl. Acad. Sci. U.S.A.}, vol. 107,
  no.~44, pp. 18\,809--18\,814, 2010.

\bibitem{amend2012positive}
J.~R. Amend, E.~Brown, N.~Rodenberg, H.~M. Jaeger, and H.~Lipson, ``A positive
  pressure universal gripper based on the jamming of granular material,''
  \emph{IEEE Trans. Robot.}, vol.~28, no.~2, pp. 341--350, 2012.

\bibitem{8794068}
S.~{Li}, J.~J. {Stampfli}, H.~J. {Xu}, E.~{Malkin}, E.~V. {Diaz}, D.~{Rus}, and
  R.~J. {Wood}, ``A vacuum-driven origami “magic-ball” soft gripper,'' in
  \emph{2019 IEEE Int Conf Robotics and Automation}, May 2019, pp. 7401--7408.

\bibitem{suresh2019spatially}
S.~A. Suresh, C.~F. Kerst, M.~R. Cutkosky, and E.~W. Hawkes, ``Spatially
  variant microstructured adhesive with one-way friction,'' \emph{Journal of
  the Royal Society Interface}, vol.~16, no. 150, p. 20180705, 2019.

\bibitem{kidambi2018modular}
N.~Kidambi, R.~L. Harne, and K.-W. Wang, ``Modular and programmable material
  systems drawing from the architecture of skeletal muscle,'' \emph{Phys. Rev.
  E}, vol.~98, no.~4, p. 043001, 2018.

\end{thebibliography}

%

%
%
%




\end{document}